\documentclass[a4paper,fleqn]{cas-sc}
\usepackage[utf8]{inputenc}
\usepackage[numbers]{natbib}
\usepackage{bm}

\def\tsc#1{\csdef{#1}{\textsc{\lowercase{#1}}\xspace}}
\tsc{WGM}
\tsc{QE}
\tsc{EP}
\tsc{PMS}
\tsc{BEC}
\tsc{DE}

\begin{document}
\let\WriteBookmarks\relax
\def\floatpagepagefraction{1}
\def\textpagefraction{.001}

\shorttitle{LSTM-PINN for population forecasting}

\shortauthors{Z.Tao et~al.}

\title{An LSTM-PINN Hybrid Method to the specific problem of population forecasting}                      
\tnotemark[1,2]


%
\author[1]{Ze Tao}[orcid=0009-0004-0202-3641]
\credit{Calculation, data analyzing, manuscript writing, review and editing}
\cormark[1]
\ead{2022000228@mails.cust.edu.cn}
\cortext[1]{Corresponding author}
\affiliation[1]{organization={Nanophotonics and Biophotonics Key Laboratory of Jilin Province, School of Physics, Changchun University of Science and Technology},
                city={Changchun},
                postcode={130022},
                country={P.R. China}}
\begin{abstract}
Deep learning has emerged as a powerful tool in scientific modeling, particularly for complex dynamical systems; however, accurately capturing age-structured population dynamics under policy-driven fertility changes remains a significant challenge due to the lack of effective integration between domain knowledge and long-term temporal dependencies. To address this issue, we propose two physics-informed deep learning frameworks—PINN and LSTM-PINN—that incorporate policy-aware fertility functions into a transport-reaction partial differential equation to simulate population evolution from 2024 to 2054. The standard PINN model enforces the governing equation and boundary conditions via collocation-based training, enabling accurate learning of underlying population dynamics and ensuring stable convergence. Building on this, the LSTM-PINN framework integrates sequential memory mechanisms to effectively capture long-range dependencies in the age-time domain, achieving robust training performance across multiple loss components. Simulation results under three distinct fertility policy scenarios—the Three-child policy, the Universal two-child policy, and the Separate two-child policy—demonstrate the models’ ability to reflect policy-sensitive demographic shifts and highlight the effectiveness of integrating domain knowledge into data-driven forecasting. This study provides a novel and extensible framework for modeling age-structured population dynamics under policy interventions, offering valuable insights for data-informed demographic forecasting and long-term policy planning in the face of emerging population challenges.
\end{abstract}


\begin{highlights}
\item Introduced two deep learning–based frameworks—Physics-Informed Neural Network (PINN) and LSTM-enhanced PINN (LSTM-PINN)—for modeling age-structured population dynamics under policy-driven fertility functions.
\item Successfully embedded age- and time-dependent fertility functions into the governing transport-reaction PDE, enabling the direct inclusion of fertility policy effects in the simulation.
\item Showed that incorporating an LSTM layer allows the framework to capture long-range temporal dependencies across age and time, yielding stable training behavior across all loss components.
\item Conducted simulations for three fertility-policy scenarios—Three-child, Universal Two-child, and Separate Two-child—and highlighted marked differences in projected age distributions, underscoring demographic sensitivity to policy design.
\end{highlights}

\begin{keywords}
Long short-term
memory \sep Population forecasting \sep Physics-informed neural network \sep Hybrid method
\end{keywords}

\maketitle

\section{Introduction}
The modeling of age-structured population dynamics evolution is of increasing significance\cite{asao2024japan} in the context of global demographic transitions and more intricate fertility policy measures. As population aging intensifies\cite{ref1} and fertility rates exhibit fluctuations\cite{ref2,chen2023changing}, precise forecasting models are critical for guiding public policy, strategizing social service allocation, and forecasting long-term economic consequences. This requires the creation of modeling frameworks that can effectively integrate biological population processes with temporal and age-dependent external factors, such as policy adjustments.

A wide range of methods have been proposed to model age-structured populations, spanning classical compartmental models\cite{bernardi2025heterogeneously,diekmann2023systematic,zhao2023comparative}, partial differential equations (PDEs)\cite{losanova2023boundary} and contemporary data-driven techniques\cite{foutel2022individual,halder2023numerical,halder2024higher}. Traditional mathematical frameworks\cite{malafeyev2024modelingdemographicproblemusing} have offered valuable insights into population dynamics under idealized conditions, while recent advancements in machine learning\cite{chen2025data} have facilitated more flexible, data-adaptive modeling strategies. However, many existing models encounter difficulties in incorporating complex temporal dependencies and policy-driven heterogeneity in a unified, interpretable, and computationally stable manner. This constrains their applicability in scenarios where long-term forecasting under policy variations is necessary.

Recent advances in physics-informed neural networks (PINNs)\cite{ali2025data,tao2025analytical} have created new opportunities for incorporating prior domain knowledge into machine learning models by embedding governing equations directly into the learning process. This approach maintains interpretability and physical consistency while utilizing the flexibility of neural networks. Extensions of PINNs, such as those incorporating recurrent architectures, including Long Short-Term Memory (LSTM) networks\cite{123}, provide enhanced capabilities for capturing long-range temporal dependencies—an essential feature in demographic systems where historical trends strongly impact future outcomes.

To promote the integration of data-driven learning with mechanistic population modeling, we implement both standard PINN and an LSTM-augmented PINN framework to simulate age-structured demographic evolution under varying fertility policies. By embedding age- and time-dependent fertility functions into a transport-reaction PDE, these models integrate domain knowledge with neural approximators, enabling the accurate representation of both biological dynamics and policy-driven population shifts. The results demonstrate stable convergence, reliable learning of long-term temporal dependencies, and interpretable predictions across policy scenarios. This work emphasizes the potential of physics-informed deep learning in advancing demographic forecasting and establishes a foundation for future extensions incorporating empirical data and policy complexity.
\section{Problem Setup}
Consider the domain $D=\left\{(a,t)\in[0,a_{0}]\times[t_{\min},t_{\max}]\mid 0\le a\le a_{0},;t_{\min}\le t\le t_{\max}\right\}$, where $a$ denotes age and $t$ denotes time. The population density function $n(a,t)$ on D satisfies the McKendrick–von Foerster equation for an age-structured demographic model:
\begin{subequations}\label{1}
	\begin{align}
		&\frac{\partial P(a, t)}{\partial t}+\alpha\frac{\partial P(a, t)}{\partial a}=-\mu(a) P(a, t) \\
		B.C.\quad&P(0, t)=\int_0^{a_{0}} b(a,t) P(a, t) \mathrm{d} a\quad\\
		I.C.\quad&P(a, 0)=P_{\mathrm{data}}(a, 0)
	\end{align}
\end{subequations}

  Where $P(a,t)$ denotes the population density at age $a$ and time $t$; $\mu(a)$ denotes the age-specific mortality rate; and $b(a,t)$ denotes the age-specific fertility rate (ASFR)
  . $\alpha=\frac{t_{max}-t_{min}}{a_{0}}$ denotes a dimensionless parameter of the "time-age scaling factor" and is denoted by and for the aging rate.
\section{Population Forecasting via PINN and LSTM-PINN Models}
\subsection{Population Dynamics Modeling with PINN}
For population forecasting, the loss function can be can be written as: 
\begin{equation}\label{2}
\mathcal{L}(\theta) = \lambda_1 \mathcal{L}_1(\theta) + \lambda_2 \mathcal{L}_2(\theta) + \lambda_3 \mathcal{L}_3(\theta),
\end{equation}
with the items: 
\begin{subequations}
\begin{align}
\mathcal{L}_1(\theta) &= \frac{1}{N} \sum_{i=1}^{N} \left( \frac{\partial P}{\partial t}(a_i, t_i) + \alpha \frac{\partial P}{\partial a}(a_i, t_i) + \mu(a_i) P(a_i, t_i) \right)^2, \\
\mathcal{L}_2(\theta) &= \frac{1}{M} \sum_{j=1}^{M} \left( \frac{P(a_j, t_{\min}) - P_0(a_j)}{P(a_j, t_{min}) + \epsilon_{0}} \right)^2, \\
\mathcal{L}_3(\theta) &= \frac{1}{K} \sum_{k=1}^{K} \left( P(0, t_k) - \int_0^{a_{0}} b(a, t_k) P(a, t_k)\, da \right)^2.
\end{align}
\end{subequations}
In the composite loss function defined in Eq.\eqref{2}, each term plays a distinct role in constraining the neural network solution. Specifically, \( \mathcal{L}_1(\theta) \) corresponds to the PDE residual loss, which enforces the population balance equation at interior sampling points; \( \mathcal{L}_2(\theta) \) represents the boundary condition loss, derived from the integral constraint at the boundary \( a = 0 \); and \( \mathcal{L}_3(\theta) \) denotes the initial condition loss, which ensures agreement with known population data at the initial time.

The variables \( a_i \), \( a_j \), and \( t_k \) denote the sampled spatial (age) and temporal points, respectively, used for computing the residual and constraints. The variable \( \epsilon_{0} \) is a small positive number introduced to stabilize the denominator in the initial condition loss and to prevent numerical divergence when the reference population density is near zero. The weighting coefficients $\lambda_{1/2/3}$ are hyperparameters that control the relative importance of the PDE residual, boundary, and initial condition terms in the total loss function. Their values are typically empirically selected to achieve balanced optimization among all loss components.

\begin{figure}[htbp]
	\centering
	\includegraphics[scale=0.3]{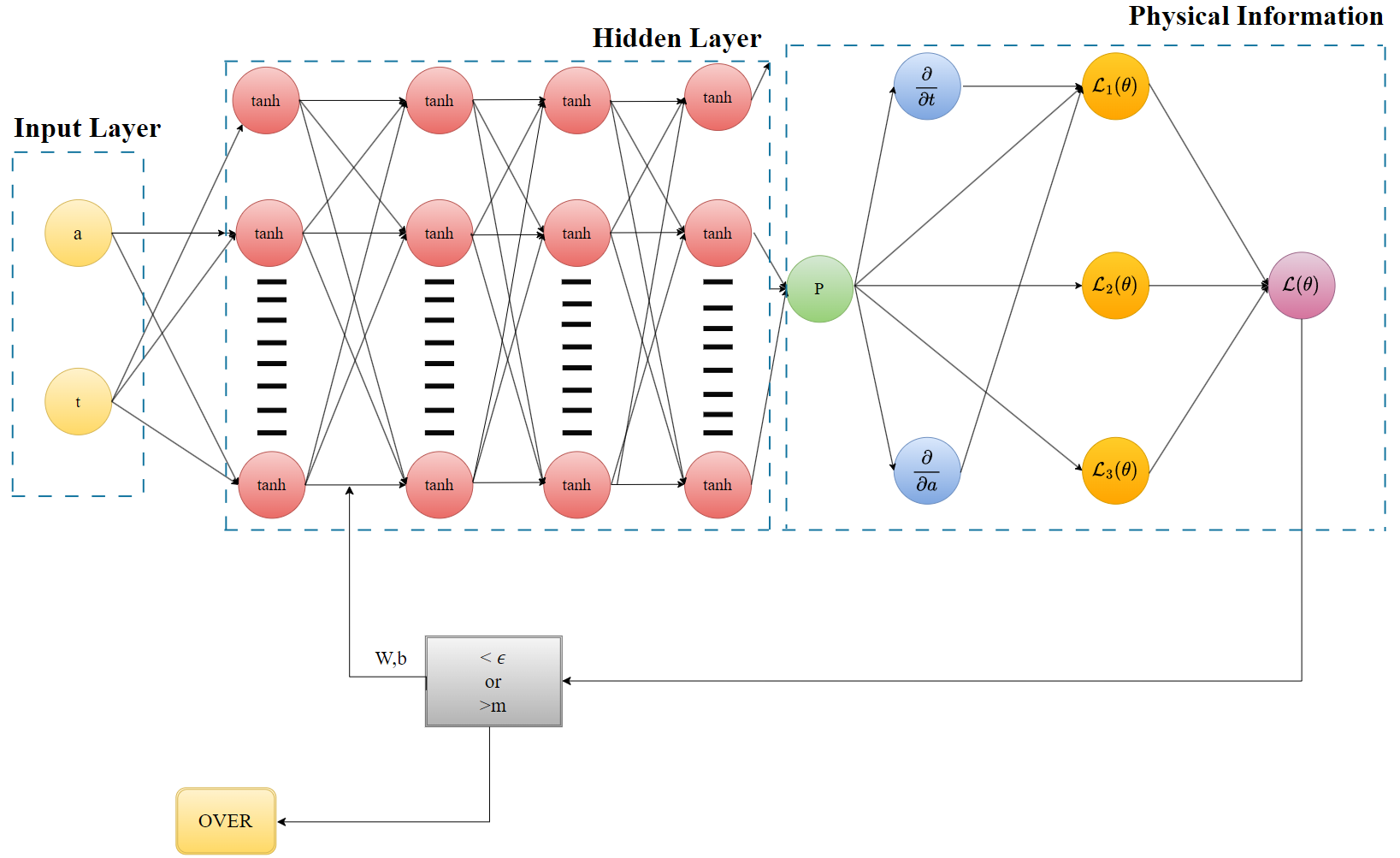}
	\caption{PINN structure diagram}
	\label{pinn}
\end{figure}

We employ a physics-informed neural network (PINN) to predict the spatiotemporal distribution of the population. As illustrated in Fig.\ref{1}, the population density function \( P(a,t) \) is initially approximated by a fully connected neural network. Spatial (age) and temporal (time) samples are fed into the network, and the residuals of the governing partial differential equation are calculated using automatic differentiation. The initial age distribution and the boundary birth condition are incorporated into the loss function as penalty terms. The total loss, comprising PDE residuals and constraint violations, is minimized using stochastic gradient descent for iterative optimization of the neural network parameters \( \theta = [W, b] \). The training process can be configured to terminate when the loss falls below a prescribed threshold \( \epsilon \), or when the number of iterations reaches a maximum value \( m \).

\subsection{Enhanced Population Prediction with LSTM-PINN}
\begin{figure}[htbp]
	\centering
	\includegraphics[scale=0.3]{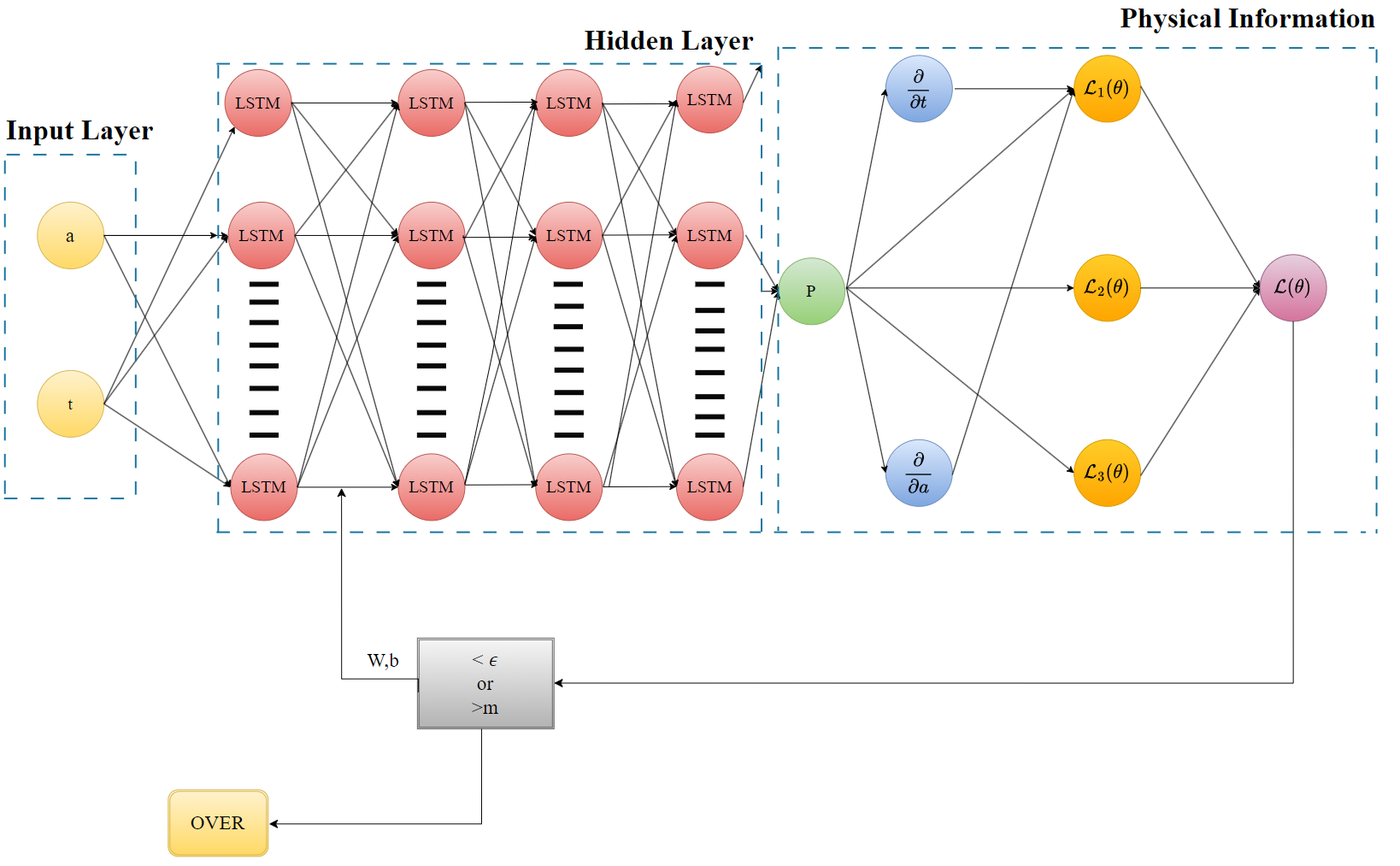}
	\caption{LSTM-PINN structure diagram}
	\label{lstm-pinn}
\end{figure}
As illustrated in Fig.\ref{lstm-pinn}, the LSTM-PINN hybrid network is employed to predict the spatiotemporal distribution of the population by integrating the capabilities of Long Short-Term Memory (LSTM) networks with Physics-Informed Neural Networks (PINN). The LSTM network consists of multiple layers that capture temporal dependencies, enabling the model to learn complex dynamics over time. The LSTM architecture processes both spatial (age) and temporal (time) input features, capturing the evolution of the population density function \( P(a,t) \) over various age groups and time points. The output of the LSTM is subsequently passed through a fully connected layer to yield the predicted population density at each spatial and temporal point.

The key advantage of the LSTM network lies in its ability to capture time-dependent processes through its gated architecture. The gates in an LSTM, including the input, forget, and output gates, enable the model to selectively retain important information over time while discarding irrelevant data, making it well-suited for capturing long-term dependencies in dynamic systems such as population dynamics. The forget gate, for instance, allows the network to disregard outdated information, while the input gate facilitates the incorporation of new and relevant information. This structure enhances the model's ability to generalize across different time scales and spatial configurations.

The physics-based constraints are incorporated into the LSTM-PINN framework through a composite loss function, which comprises terms corresponding to the residuals of the governing population balance equation, as well as the boundary and initial conditions. These constraints ensure that the model adheres to the physical laws governing the system, thereby enforcing the population balance equation while simultaneously exploiting the temporal modeling capabilities of the LSTM. The total loss is minimized using stochastic gradient descent (SGD), which iteratively updates the network's parameters \( \theta = [W, b] \) through backpropagation of gradients obtained via automatic differentiation.

Similar to the traditional PINN framework, the training process terminates when the loss function falls below a predefined threshold \( \epsilon \) or when the number of iterations exceeds a set limit \( m \). This approach allows the model to efficiently predict the population distribution over time while ensuring compliance with the governing physical constraints.

\section{Numerical Examples and Solutions}
\subsection{Model Formulation and Specific Problem Definition}
Set the value of \( a_{0} = 100 \), with \( t_{\text{min}} = 2024 \) and \( t_{\text{max}} = 2054 \), respectively. Additionally, \( \mu(a) \) is defined as:
$$
\mu(a)= \begin{cases}\mu_0+B a, & 0 \leq a<60, \\ \left(\mu_0+B \cdot 60\right) \exp [0.06(a-60)], & a \geq 60,\end{cases}
$$
where
$\mu_{0}=0.006805083$, $B=0.0003$.

The age-time dependent fertility rate function \( b(a,t) \) varies according to the policy implemented. Under the three-child policy, we define it as:
$$
b(a, t)=\min \left\{\operatorname{base} \_\operatorname{asfr}(a) \cdot\left[1+0.2 \cdot \mathbf{1}_{t \geq 2014}+0.2 \cdot \mathbf{1}_{t \geq 2016}+0.2 \cdot \mathbf{1}_{t \geq 2021}\right], 0.25\right\};
$$

Under the two-child policy, we define it as:
$$
b(a, t)=\min \left\{\operatorname{base} \_\operatorname{asfr}(a) \cdot\left[1+0.2 \cdot \mathbf{1}_{t \geq 2024}\right], 0.20\right\};
$$

Under the universal two-child policy, we define it as: $$
b(a, t)=\min \left\{\operatorname{base} \_\operatorname{asfr}(a) \cdot\left[1+0.2 \cdot \mathbf{1}_{t \geq 2024}\right], 0.25\right\};
$$

Where the indicator function $\mathbf{1}_{t \geq t_{1}}$ is defined as: 
$$
\mathbf{1}_{t \geq t_{1}}= \begin{cases}1, & t \geq t_{1} \\ 0, & t<t_{1}\end{cases};
$$
And age appropriate fertility rate$\operatorname{base} \_\operatorname{asfr}(a) $ is defined as: 
\begin{equation*}
\text{base\_asfr}(a) = \begin{cases} 
0.0022 \cdot (a - 20)(35 - a), & 20 \leq a \leq 35 \\
0, & \text{otherwise}
\end{cases};
\end{equation*}
\subsection{PINN Approach to Numerical Solutions}
The Physics-Informed Neural Network (PINN) employed in this study consists of a fully connected feedforward neural network with four hidden layers. The input consists of two normalized variables—age and time—while the hidden layers contain 128, 128, and 64 neurons respectively, each activated by the hyperbolic tangent (Tanh) function. The output layer yields a single value representing the normalized population density. The model is trained using the Adam optimizer with a fixed learning rate of \(5 \times 10^{-4}\), with training conducted over 10,000 epochs. In each epoch, \(N = 5,000\) collocation points are used to enforce the partial differential equation (PDE), \(M = 2,000\) points are used for initial conditions at \(t = 2024\), and \(K = 2,000\) points are employed for boundary conditions at age \(a = 0\). These sampling strategies are repeated in each epoch to enhance generalization and stability.
\begin{figure}[htbp]
	\centering
	\includegraphics[scale=0.3]{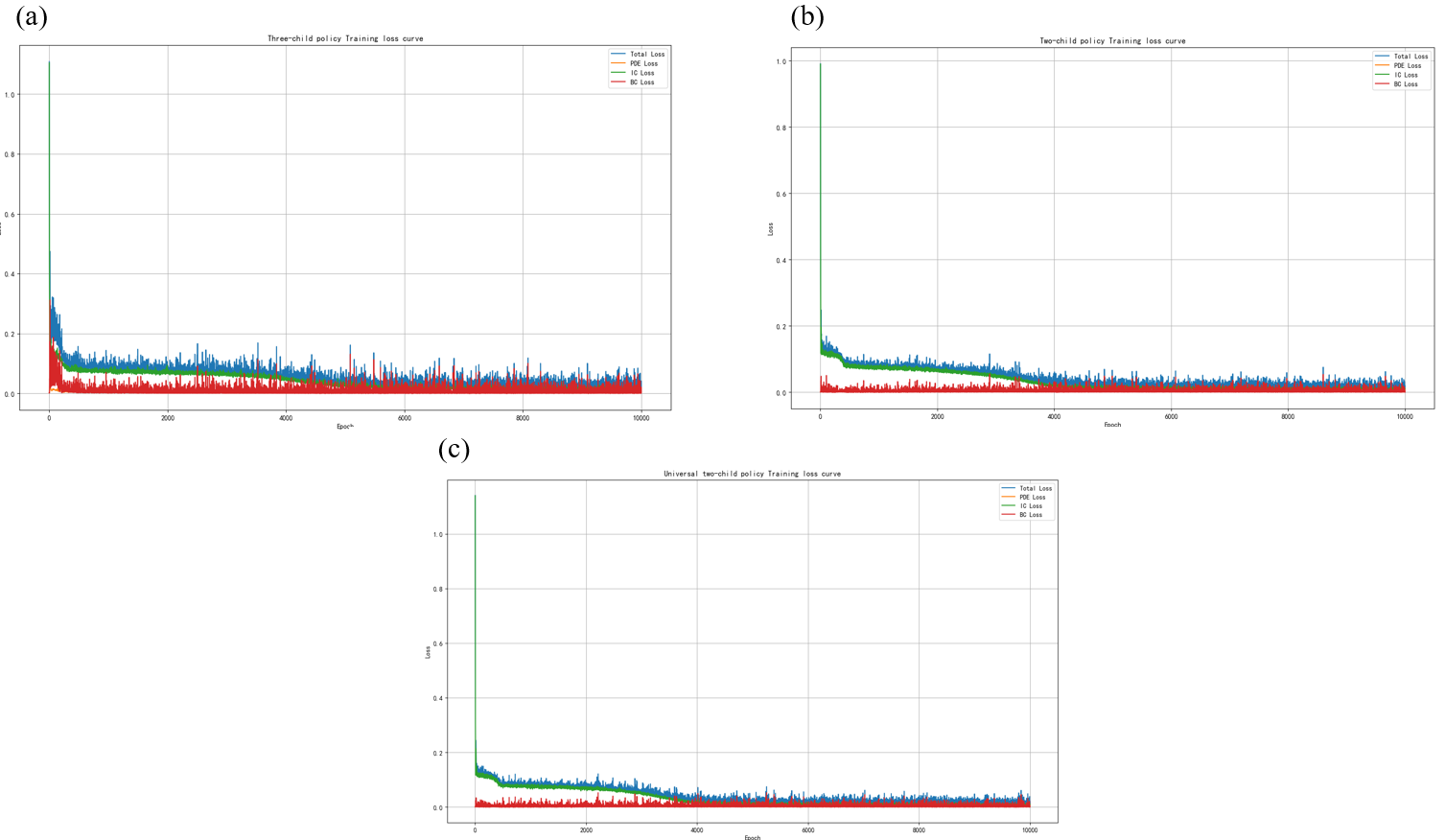}
	\caption{The loss function of PINN with different policy}
	\label{LOSS for PINN}
\end{figure} 
\begin{figure}[htbp]
	\centering
	\includegraphics[scale=0.3]{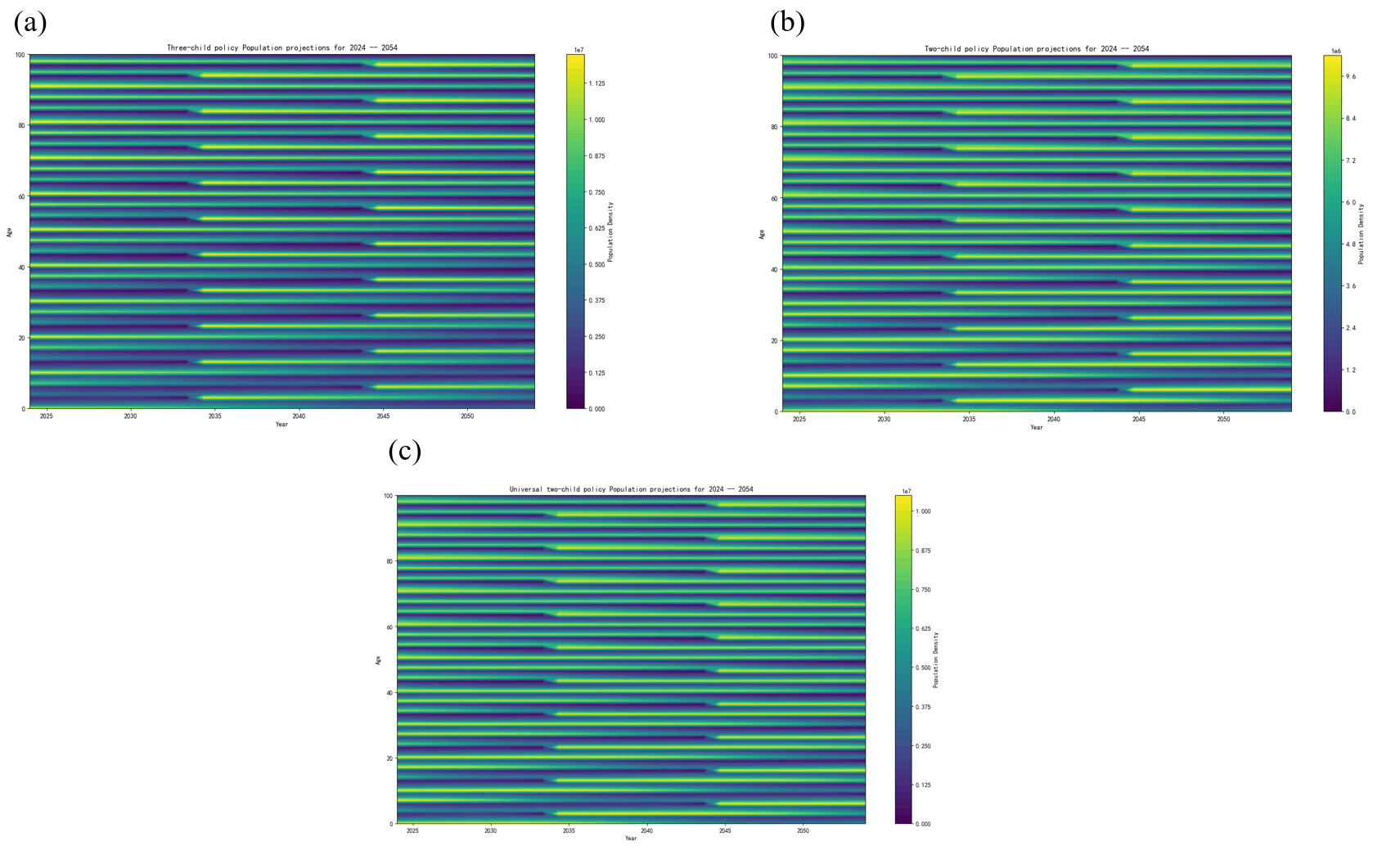}
	\caption{PINN results with different policy}
	\label{Results for PINN}
\end{figure} 
The loss function results are shown in Fig.\ref{LOSSforPINN}, where the total loss, PDE residual loss, initial condition loss, and boundary condition loss all exhibit steady convergence throughout training, demonstrating the model's ability to capture the underlying dynamics. The corresponding population projection results for three distinct fertility policy scenarios—the "Three-child policy", the "Two-child policy", and the "Universal two-child policy"—are illustrated in Fig.\ref{ResultsforPINN}, highlighting notable variations in age-time demographic distributions under each policy over the forecast period from 2024 to 2054.
\subsection{LSTM-PINN Approach to Numerical Solutions}
\begin{figure}[htbp]
	\centering
	\includegraphics[scale=0.3]{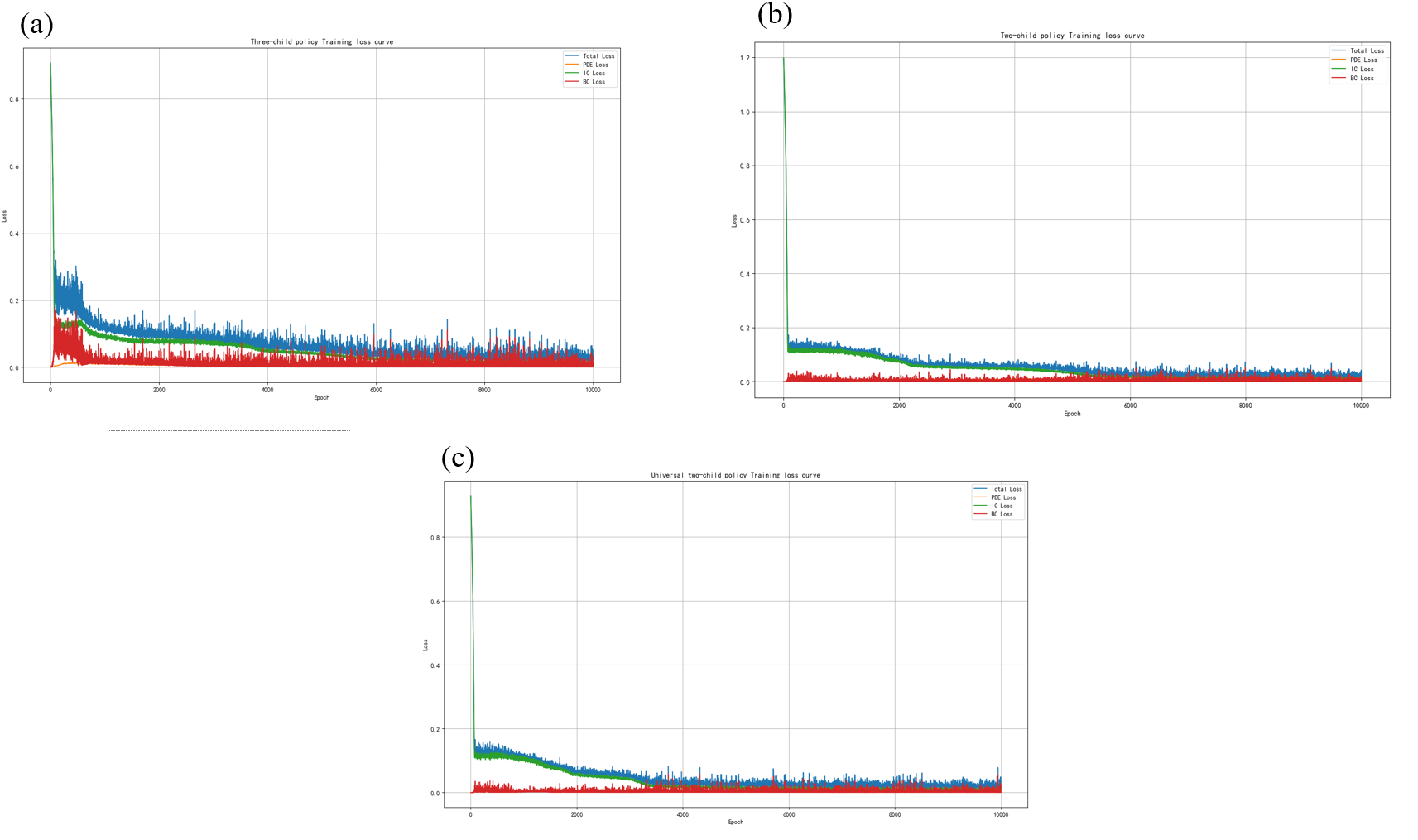}
	\caption{The loss function of PINN with different policy}
	\label{LOSS for LSTM-PINN}
\end{figure} 
\begin{figure}[htbp]
	\centering
	\includegraphics[scale=0.3]{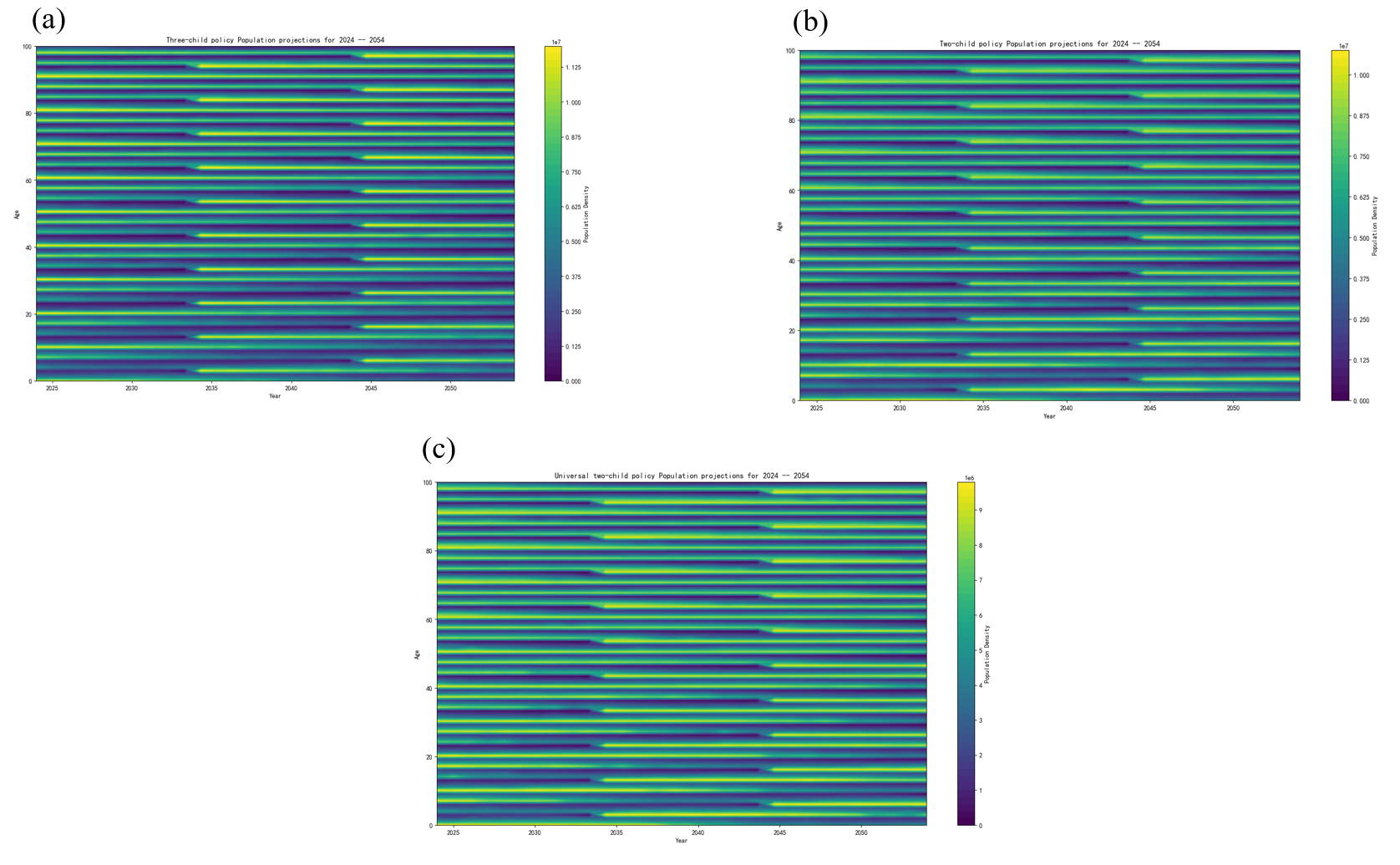}
	\caption{LSTM-PINN results with different policy}
	\label{Results for LSTM-PINN}
\end{figure} 
In our LSTM-PINN framework, the neural network model employs a stacked LSTM architecture with 4 layers, each consisting of 64 units. The input to the LSTM at each time step is a two-dimensional vector. Internally, each LSTM unit employs the canonical gated structure, which consists of three types of gates per unit: the input gate, the forget gate, and the output gate. These gates control the information flow across time steps, enabling the model to capture nonlinear dynamics and long-range dependencies in the age-time domain. With 4 layers, each containing 64 units, the total number of gates within the LSTM block amounts to 768 gates (i.e., 4 × 64 × 3). Each gate is implemented via a learnable affine transformation followed by a sigmoid activation, regulating cell state updates and memory retention. The network employs a dropout rate of 0.1 between LSTM layers to mitigate overfitting. After the LSTM layers, the output from the final time step is passed through a fully connected linear layer to predict the normalized population density.

During training, we use the Adam optimizer with a fixed learning rate of \(5 \times 10^{-4}\), and the model is trained for 10,000 epochs per scenario. Each epoch employs a sampling strategy consisting of \(N = 5,000\) interior points to enforce the PDE constraint, \(M = 2,000\) initial condition points to fit the initial age distribution at the starting year, and \(K = 2,000\) boundary condition points to handle the birth-related integral boundary constraint.

The loss function is a composite of three components: a residual loss enforcing the transport-reaction PDE, an initial condition loss ensuring consistency with the known population at the initial time, and a boundary condition loss enforcing a time-dependent birth integral constraint. Each component is equally weighted in the total loss. As shown in Fig.\ref{LOSS for LSTM-PINN}, the loss function demonstrates stable convergence behavior, with all three components—PDE, IC, and BC—decreasing gradually, contributing to a steady reduction in total loss across training epochs. The predicted population density distributions from 2024 to 2054 under three policy scenarios—Three-child policy, Two-child policy, and Universal two-child policy—are presented in Fig.\ref{Results for LSTM-PINN}, highlighting notable variations in age-time dynamics and population growth trajectories driven by varying fertility policies.
\section{Conclusion}
In this study, we developed and evaluated two deep learning-based frameworks—PINN and LSTM-PINN—for simulating the evolution of age-structured population density under different fertility policy scenarios spanning the period from 2024 to 2054. By embedding policy-driven, age- and time-dependent fertility functions into the governing transport-reaction PDE, the models successfully model both biological processes and the demographic shifts induced by policy changes.

The standard PINN model demonstrated stable convergence and accurately learned the underlying population dynamics by enforcing the governing equation and associated conditions through collocation-based methods. The LSTM-PINN framework, incorporating sequential memory mechanisms, effectively captured long-range temporal dependencies within the age-time domain and exhibited stable training behavior across all loss components.

Simulation results across three distinct policy scenarios—the Three-child policy, the Universal two-child policy, and the Separate two-child policy—revealed substantial differences in the projected population distributions, reflecting the sensitivity of demographic dynamics to fertility policy. Both models offer effective methodologies for integrating domain knowledge into data-driven demographic forecasting tasks.

While the current study provides a preliminary demonstration of using PINN and LSTM-PINN for age-structured population modeling under different fertility policies, it constitutes a preliminary investigation. Future work will involve a more comprehensive analysis, including systematic experiments under varying learning rates for both methods to evaluate stability and performance variations. Potential algorithmic improvements and the integration of real demographic data will also be explored to enhance the realism and policy relevance of the simulations, particularly in the context of population challenges faced by China. Furthermore, more rigorous mathematical formulations and refined modeling assumptions are anticipated to produce more accurate and interpretable results.

The complete source code for both methods has been made publicly available on GitHub (see the "Data Availability" section below), and we encourage contributions from the research community to further improve and extend this work.

\printcredits
\section*{Data availability}
The dataset and codes generated and/or analyzed during the current study are available at GitHub : \\https://github.com/Uderwood-TZ/LSTM-PINN-and-PINN-for-population-forecasting.git 
\bibliographystyle{unsrt}
\bibliography{cas-refs}

\end{document}